\documentclass{article}

\usepackage{amsmath,amssymb,amsthm}
\usepackage{bm}

\usepackage{times}
\usepackage{graphicx} 
\usepackage{subfigure} 

\usepackage{natbib}

\usepackage{algorithm}
\usepackage{algorithmic}

\usepackage{hyperref}

\usepackage[accepted]{icml2016}

\renewcommand{\algorithmicrequire}{\textbf{Input:}}
\renewcommand{\algorithmicensure}{\textbf{Output:}}
\renewcommand\algorithmiccomment[1]{{\small/*{\it#1}*/}}

\theoremstyle{plain}
\newtheorem{thm}{Theorem}[section]

\theoremstyle{definition}
\newtheorem{defn}{Definition}[section]
\newtheorem{assum}{Assumption}[section]

\theoremstyle{remark}

\icmltitlerunning{Learning Granger Causality for Hawkes Processes}

\begin{document} 

\twocolumn[
\icmltitle{Learning Granger Causality for Hawkes Processes}

\icmlauthor{Hongteng Xu}{hxu42@gatech.edu}
\icmladdress{School of ECE, Georgia Institute of Technology}
\icmlauthor{Mehrdad Farajtabar}{mehrdad@gatech.edu}
\icmladdress{College of Computing, Georgia Institute of Technology}
\icmlauthor{Hongyuan Zha}{zha@cc.gatech.edu}
\icmladdress{College of Computing, Georgia Institute of Technology}

\icmlkeywords{Hawkes processes, impact functions, Granger causality, sparse-group-lasso}

\vskip 0.3in
]

\begin{abstract} 
Learning Granger causality for general point processes is a very challenging task. 
In this paper, we propose an effective method, \emph{learning Granger causality}, for a special but significant type of point processes --- Hawkes process.
According to the relationship between Hawkes process's impact function and its Granger causality graph, our model represents impact functions using a series of basis functions and recovers the Granger causality graph via group sparsity of the impact functions' coefficients. 
We propose an effective learning algorithm combining a maximum likelihood estimator (MLE) with a sparse-group-lasso (SGL) regularizer. 
Additionally, the flexibility of our model allows to incorporate the clustering structure event types into learning framework.
We analyze our learning algorithm and propose an adaptive procedure to select basis functions.
Experiments on both synthetic and real-world data show that our method can learn the Granger causality graph and the triggering patterns of the Hawkes processes simultaneously. 
\end{abstract} 

\section{Introduction}\label{intro}
In many practical situations, we need to deal with a large amount of irregular and asynchronous sequential data observed in continuous time. The applications include the user viewing records in an IPTV system (when and which TV programs are viewed), and the patient records in hospitals (when and what diagnoses and treatments are given), among many others.
All of these data can be viewed as event sequences containing multiple event types and modeled via multi-dimensional point processes. 
A significant task for a multi-dimensional point process is to learn the so-called Granger causality.
From the viewpoint of graphical models, it means to construct a directed graph called Granger causality graph (or local independence graph)~\cite{didelez2008graphical} over the dimensions (i.e., the event types) of the process. 
The arrow connecting two nodes indicates that the event of the dimension corresponding to the destination node is 
dependent on the historical events of the dimension corresponding to the source node. 
Learning Granger causality for multi-dimensional point processes is meaningful for many practical applications. 
Take our previous two examples: 
the Granger causality among IPTV programs reflects users' viewing preferences and patterns, which is important for personalized program recommendation and IPTV system simulation; 
the Granger causality among diseases helps us to construct a disease network, which is beneficial to predict potential diseases for patients and leads to more effective treatments. 

Unfortunately, learning Granger causality for general multi-dimensional point processes is very challenging. 
Existing works mainly focus on learning Granger causality for time series~\cite{arnold2007temporal,eichler2012graphical,basu2015network}, where the Granger causality is captured via the so-called vector auto-regressive (VAR) model~\cite{han2013transition} based on  discrete time-lagged variables. 
For point processes, on the contrary, the event sequence is in continuous time and no fixed time-lagged observation is available.
Therefore, it is hard to find a universal and tractable representation of the 
complicated historical events to describe Granger causality for the process. 
A potential solution is to construct features for various dimensions from historical events and learn Granger causality via feature selection~\cite{lianmultitask}. 
However, this method is highly dependent on the specific feature construction method used, resulting in dubious  Granger causality.
 
To make concrete progress, we focus on a special class of point processes called Hawkes processes and their Granger causality. 
Hawkes processes are widely used and are capable of describing the self-and mutually-triggering patterns among different event types. 
Applications include bioinformatics~\cite{reynaud2010adaptive}, social network analysis~\cite{zhao2015seismic}, financial analysis~\cite{bacry2013some}, etc. 
Learning Granger causality will further extend applications of Hawkes processes in many other fields.

Technically, based on the graphical model of point process~\cite{didelez2008graphical}, the Granger causality of Hawkes process can be captured by its impact functions. 
Inspired by this fact, we propose a nonparametric model of Hawkes processes, where the impact functions are represented by a series of basis functions, and we discover the Granger causality via group sparsity of impact functions' coefficients.   
Based on the explicit representation of Granger causality, we propose a novel learning algorithm combining the maximum likelihood estimator with the sparse-group-lasso (SGL) regularizer on impact functions. 
The pairwise similarity between various impact functions is considered when the clustering structure of event types is available. 
Introducing these structural constraints enhances the robustness of our method.
The learning algorithm applies the EM-based strategy~\cite{lewis2011nonparametric,zhou2013learning} and obtains close-form solutions to update model's parameters iteratively. 
Furthermore, we discuss the selection of basis function based on sampling theory, and provide a useful guidance for model selection. 

Our method captures Granger causality from complicated event sequences in continuous time. 
Compared with existing learning methods for Hawkes processes~\cite{zhouke2013learning,eichler2015graphical}, our model avoids discretized representation of impact functions and conditional intensity, and considers the induced structures across impact functions. 
These improvements not only reduce the complexity of the learning algorithm but also improve learning performance. 
We investigate the robustness of our method to the changes of parameters and test our method on both synthetic and real-world data. 
Experimental results show that our method can indeed reveal the Granger causality of Hawkes processes and obtain superior learning performance compared with other competitors.

\section{Related Work}\label{relate} 
\textbf{Granger causality.} Many efforts have been made to learn the Granger causality of point processes~\cite{meek2014toward}.  
For general random processes, a kernel independence test is developed in~\cite{chwialkowski2014kernel}. 
Focusing on 1-D point process with simple piecewise constant conditional intensity, a model for capturing temporal dependencies between event types is proposed in~\cite{gunawardana2011model}. 
In~\cite{basu2015network,song2013identification}, the inherent grouping structure is considered when learning the Granger casuality on networks from discrete transition process. 
\cite{daneshmand2014estimating} proposed a continuous-time diffusion network inference method based on parametric cascade generative process. 
In more general cases, a class of graphical models of marked point processes is proposed in~\cite{didelez2008graphical} to capture the local independence over various marks.
Specializing the work for Hawkes processes, \cite{eichler2015graphical} firstly connects Granger causality with impact functions. 
However, although applying lasso or its variants to capture the intra-structure of nodes~\cite{ahmed2009recovering} is a common strategy, less work has been done on learning causality graph of Hawkes process with sparse-group-lasso as we do, which leads them to be sensitive to noisy and insufficient data.

\textbf{Hawkes processes.} Hawkes processes~\cite{hawkes1971spectra} are proposed to model complicated event sequences where historical events have influences on future ones. 
It is applied to many problems, e.g., seismic analysis~\cite{daley2007introduction}, financial analysis~\cite{bacry2013some}, social network modeling~\cite{FarajtabarCoevolutionNIPS15,zhou2013learning,zhouke2013learning} and bioinformatics~\cite{reynaud2010adaptive,carstensen2010multivariate}. 
Most of existing works use predefined impact function with known parameters, e.g., the exponential functions in~\cite{farajtabar2014shaping, rasmussen2013bayesian,zhou2013learning,hall2014tracking,yan2015machine} and the power-law functions in~\cite{zhao2015seismic}. 
For enhancing the flexibility, a nonparametric model of 1-D Hawkes process is first proposed in~\cite{lewis2011nonparametric} based on ordinary differential equation (ODE) and extended to multi-dimensional case in~\cite{zhouke2013learning,luo2015multi}. 
Similarly, \cite{bacry2012non} proposes a nonparametric estimation of Hawkes processes via solving the Wiener-Hopf equation. 
Another nonparametric strategy is the contrast function-based estimation in~\cite{reynaud2010adaptive,hansen2015lasso}. 
It minimizes the estimation error of conditional intensity function and leads to a Least-Squares (LS) problem~\cite{eichler2015graphical}. 
\cite{du2012learning,lemonnier2014nonparametric} decompose impact functions into basis functions to avoid discretization. 
The Gaussian process-based methods~\cite{adams2009tractable,lloyd2015variational,lianmultitask,samo2015scalable} have been reported to successfully estimate more general point processes. 

\section{Basic Concepts}
%
\subsection{Temporal Point Processes}
A temporal point process is a random process whose realization consists of a list of discrete events in time $\{ t_i \}$ with $t_i \in [0,T]$. 
Here $[0,T]$ is the time interval of the process.
It can be equivalently represented as a counting process,
$N=\{N(t)|t\in [0,T]\}$, where $N(t)$ records the number of events before time $t$.
A multi-dimensional point process with $U$ types of event is represented by $U$ counting processes $\{N_u\}_{u=1}^{U}$ on a probability space $(\Omega, \mathfrak{F}, \mathbb{P})$. 
$N_u=\{N_u(t)| t\in [0,T]\}$, where $N_u(t)$ is the number of type-$u$ events occurring at or before time $t$. 
$\Omega=[0,T]\times \mathcal{U}$ is the sample space. 
$\mathcal{U}=\{1,...,U\}$ is the set of event types. 
$\mathfrak{F}=(\mathfrak{F}(t))_{t\in\mathbb{R}}$ is the filtration representing the set of events sequence the process can realize until time $t$. 
$\mathbb{P}$ is the probability measure.
A way to characterize point processes is via the conditional intensity function capturing the patterns of interests, i.e., self-triggering or self-correcting~\cite{xu2015trailer}. 
It is defined as the expected instantaneous rate of happening type-$u$ events given the history:
\begin{eqnarray*}
\begin{aligned}
\lambda_u(t)dt = \lambda_u(t|\mathcal{H}_t^{\mathcal{U}})dt=\mathbb{E}[dN_u(t)|\mathfrak{F}(t)].
\end{aligned}
\end{eqnarray*}
Here $\mathcal{H}_t^{\mathcal{U}}= \{(t_i, u_i) |  t_i < t, u_i\in\mathcal{U}\}$ collects historical events of all types before time $t$. 

{\bf Hawkes Processes.} A multi-dimensional Hawkes process is a counting process who has a particular form of intensity:
\begin{eqnarray}\label{intensity}
\begin{aligned}
\lambda_u(t)=\mu_u + \sideset{}{_{u'=1}^U}\sum \int_{0}^t \phi_{uu'}(s) dN_{u'}(t-s),
\end{aligned}
\end{eqnarray}
where $\mu_u$ is the exogenous base intensity independent of the history while $\sum_{u'=1}^U\int_{0}^t \phi_{uu'}(s) dN_{u'}(t-s)$ the endogenous intensity capturing the peer influence~\cite{farajtabar2014shaping}. 
Function $\phi_{uu'}(t)\geq 0$ is called impact function, which measures decay in the influence of historical type-$u'$ events on the subsequent type-$u$ events.

\subsection{Granger Causality for Point processes}
We are interested in identifying, if possible, a subset of the event types $\mathcal{V} \subset \mathcal{U}$ for the type-$u$ event, such that $\lambda_u(t)$ only depends on historical events of  types in $\mathcal{V}$, denoted as $\mathcal{H}_t^{\mathcal{V}}$, and not those of the rest types, denoted as $\mathcal{H}_t^{\mathcal{U}\setminus\mathcal{V} }$. 
From the viewpoint of graphical model, it is about local independence over the dimensions of the point process --- the occurrence of historical events in $\mathcal{V}$ influences the probability of occurrence of type-$u$ events at present and future while the occurrence of historical events in $\mathcal{U}\setminus\mathcal{V}$ does not. 
In order to proceed formally we introduce some notations.
For a subset $\mathcal{V} \subset \mathcal{U}$, let $N_{\mathcal{V}} = \{N_u(t) | u \in \mathcal{V}\}$.
The filtration $\mathfrak{F}_t^{\mathcal{V}}$ is defined as $\sigma\{ N_u(s) | s \leq t, u \in \mathcal{V}\}$, i.e., the smallest $\sigma$-algebra generated by the random processes. 
In particular, $\mathfrak{F}_t^{u}$ is the internal filtration of the counting process $N_u(t)$ while $\mathfrak{F}_t^{-u}$ is the filtration for the subset $\mathcal{U}\setminus\{u\}$.

\begin{defn}\label{Def1}
\emph{\cite{didelez2008graphical}. 
The counting process $N_u$ is locally independent of $N_{u'}$ given $N_{\mathcal{U}\setminus\{u, u'\}}$ if
the intensity function $\lambda_u(t)$ is measurable with respect to $\mathfrak{F}_t^{-u'}$ for all $t \in [0, T]$. 
Otherwise $N_u$ is locally dependent of $N_{u'}$.}
\end{defn}

Intuitively, the above definition says that $\{N_{u'}(s)| s<t\}$ does not influence $\lambda_u(t)$,
given $\{N_{l}(s)| s<t,~l \neq u'\}$. 
In \cite{eichler2015graphical}, the notion of Granger non-causality is used, and the above definition is equivalent to saying that type-$u'$ event does not Granger-cause type-$u$ event w.r.t. $\mathfrak{F}_t^{\mathcal{U}}$. 
Otherwise, we say type-$u'$ event Granger-causes type-$u$ event w.r.t. $\mathfrak{F}_t^{\mathcal{U}}$.
With this definition, we can construct the so-called \emph{Granger causality graph} $G=(\mathcal{U}, \mathcal{E})$ with the event types $\mathcal{U}$ (the dimensions of the point process) as the nodes and the directed edges indicating the causation, i.e., $u'\rightarrow u \in \mathcal{E}$ if type-$u'$ event Granger-causes type-$u$ one.

Learning Granger causality for a general multi-dimensional point process is a difficult problem. In the next section we introduce an efficient method for learning the Granger causality of the Hawkes process.

\section{Proposed Model and Learning Algorithm}
In this section, we first generalize a known result for Hawkes process. 
Then, we propose a model of Hawkes process representing impact functions via a series of basis functions. 
An efficient learning algorithm combining the MLE with the sparse-group-lasso is applied and analyzed in details. 
Compared with existing learning algorithms, our algorithm is based on convex 
optimization and has lower complexity, which learns Granger causality robustly. 

\subsection{Granger Causality of Hawkes Process}
The work in~\cite{eichler2015graphical} reveals the relationship between Hawkes processes' impact function and its Granger causality graph as follows,
\begin{thm}\label{the1}
\cite{eichler2015graphical}. 
Assume a Hawkes process with conditional intensity function defined in~(\ref{intensity}) and Granger causality graph $G(\mathcal{U}, \mathcal{E})$.  If the condition $dN_{u'}(t-s)>0$ for $0\leq s<t\leq T$ holds, then, $u'\rightarrow u \notin \mathcal{E}$ if and only if $\phi_{uu'}(t) =0$ for $t \in [0, \infty]$.
\end{thm}
In practice, Theorem~\ref{the1} can be easily specified in the time interval $t\in [0,T]$.
It provides an explicit representation of the Granger causality of multi-dimensional Hawkes process --- learning whether type-$u'$ event Granger-causes type-$u$ event or not is equivalent to detecting whether the impact function $\phi_{uu'}(t)$ is all-zero or not. 
In other words, the group sparsity of impact functions along the time dimension indicates the Granger causality graph over the dimensions of Hawkes process. 
Therefore, for multi-dimensional Hawkes process, we can learn its Granger causality via learning its impact functions, which requires tractable and flexible representations of the functions.

\subsection{Learning Task}
When we parameterize $\phi_{uu'}(t) = a_{uu'} \kappa(t)$ as~\cite{zhou2013learning} does, where $\kappa(t)$ models time-decay of event's influence and $a_{uu'} \geq 0$ captures the influence of $u'$-type events on $u$-type ones, the binarized \emph{infectivity matrix} $A=[\mbox{sign}(a_{uu'})]$ is the adjacency matrix of the corresponding Granger causality graph. 
Although such a parametric model simplifies the representation of impact function and reduces the complexity of the model, this achievement comes with the cost of inflexibility of the model --- the model estimation will be poor if the data does not conform to the assumptions of the model. 
To address this problem, we propose a nonparametric model of Hawkes processes, representing the impact function in~(\ref{intensity}) via a linear combination of basis functions as
\begin{eqnarray}\label{represent}
\begin{aligned}
\phi_{uu'}(t) = \sideset{}{_{m=1}^{M}}\sum a_{uu'}^{m}\kappa_m(t).
\end{aligned}
\end{eqnarray}
Here $\kappa_m(t)$ is the $m$-th basis function and $a_{uu'}^{m}$ is the coefficient corresponding to $\kappa_m(t)$. 
The selection of bases will be discussed later in the paper.

Suppose we have a set of event sequences $\mathcal{S}=\{s_c\}_{c=1}^{C}$. 
$s_c=\{(t_i^c, u_i^c)\}_{i=1}^{N_c}$, where $t_i^c$ is the time stamp of the $i$-th event of $s_c$ and $u_i^c\in \{1,...,U\}$ is the type of the event. 
Thus, the log-likelihood of model parameters  $\Theta=\{\bm{A}=[a_{uu'}^{m}]\in\mathbb{R}^{U\times U\times M},\bm{\mu}=[\mu_{u}]\in\mathbb{R}^{U}\}$ can be expressed as:
\begin{eqnarray}\label{loglike}
\begin{aligned}
\mathcal{L}_{\Theta}&= \sum_{c=1}^{C}\biggl\lbrace \sum_{i=1}^{N_c}\log\lambda_{u_i^c}(t_i^c) - \sum_{u=1}^{U}\int_{0}^{T_c}\lambda_u(s)ds \biggr\rbrace\\
&=\sum_{c=1}^{C}\biggl\lbrace \sum_{i=1}^{N_c}\log\biggl(\mu_{u_i^c}+\sum_{j=1}^{i-1}\sum_{m=1}^{M}a_{u_i^c u_j^c}^{m}\kappa_m(\tau_{ij}^c)\biggr)\\
&\quad -\sum_{u=1}^{U}\biggl( T_c\mu_u + \sum_{i=1}^{N_c}\sum_{m=1}^{M}a_{uu_i^c}^{m}K_m(T_c-t_i^c)\biggr) \biggr\rbrace,
\end{aligned}
\end{eqnarray}
where $\tau_{ij}^c=t_i^c-t_j^c$, $K_m(t)=\int_{0}^{t}\kappa_m(s)ds$. 
For constructing Granger causality accurately and robustly, we consider the following three types of regularizers:

\textbf{Local Independence.} According to Theorem~\ref{the1}, the $u'$-type event has no influence on the $u$-type one (i.e., directed edge $u' \rightarrow u \notin \mathcal{E}$) if and only if $\phi_{uu'}(t)=0$ for all $t\in \mathbb{R}$, which requires $a_{uu'}^m=0$ for all $m$. 
Therefore, we use group-lasso~\cite{yang2010online,simon2013sparse,song2013identification} to regularize the coefficients of impact functions, denoted as $\|\bm{A}\|_{1,2}=\sum_{u,u'}\|a_{uu'}\|_2$, where $a_{uu'}=[a_{uu'}^1,...,a_{uu'}^M]^{\top}$. 
It means that along the time dimension the coefficients' tensor $\bm{A}$ should yield to the constraint of group sparsity. 

\textbf{Temporal Sparsity.} A necessary condition for the stationarity of Hawkes process is  $\int_{0}^{\infty}\phi_{ij}(s)ds<\infty$, which means $\lim_{t\rightarrow \infty}\phi_{ij}(t)\rightarrow 0$. 
Therefore, we add sparsity constraints to the coefficients of impact functions, denoted as $\|\bm{A}\|_1=\sum_{u,u',m}|a_{uu'}^m|$.

\textbf{Pairwise Similarity.} Event types of Hawkes process may exhibit clustering structure. For example, if $u$ and $u'$ are similar event types, their influences on other event types should be similar (i.e., $\phi_{\cdot u}(t)$ are close to $\phi_{\cdot u'}(t)$) and the influences of other event types on them should be similar as well (i.e., $\phi_{u\cdot}(t)$ are close to $\phi_{u'\cdot}(t)$). 
When the clustering structure is (partially) available, we add constraints of pairwise similarity on the coefficients of corresponding impact functions as follows
\begin{eqnarray*}
\label{pair}
\begin{aligned}
E(\bm{A})=\sideset{}{_{u=1}^{U}}\sum\sideset{}{_{u'\in\mathcal{C}_u}}\sum\|a_{u\cdot}-a_{u'\cdot}\|_F^2+\|a_{\cdot u'}-a_{\cdot u}\|_F^2.
\end{aligned}
\end{eqnarray*}
$\mathcal{C}_u$ contains the event types within the cluster that the event of $u$ type resides. 
$a_{u\cdot}\in\mathbb{R}^{U\times M}$ is the slice of $\bm{A}$ with row index $u$, and $a_{\cdot u}\in\mathbb{R}^{U\times M}$ is the slice with column index $u$. 
In summary, the learning problem of the Hawkes process is
\begin{eqnarray}\label{obj}
\begin{aligned}
\min_{\Theta\geq\bm{0}}&-\mathcal{L}_{\Theta}+\alpha_{S}\|\bm{A}\|_1
+\alpha_{G}\|\bm{A}\|_{1,2}+\alpha_{P}E(\bm{A}).
\end{aligned}
\end{eqnarray}
Here $\alpha_{S}$, $\alpha_G$ and $\alpha_P$ control the influences of the regularizers. 
The nonnegative constraint guarantees the model being physically-meaningful.

\subsection{An EM-based Algorithm}
Following~\cite{lewis2011nonparametric,zhouke2013learning}, we propose an EM-based learning algorithm for solving optimization problem~(\ref{obj}) iteratively. 
Specifically, given current parameters $\Theta^{(k)}$, we first apply the Jensen's inequality and construct a tight upper-bound of log-likelihood function appeared in~(\ref{loglike}) as follows:
\begin{eqnarray*}\label{surrogate}
\begin{aligned}
&Q_{\Theta}^{(k)}=\sum_{c=1}^{C}\biggl\lbrace -\sum_{u=1}^{U}\biggl( T_c\mu_u + \sum_{i=1}^{N_c}\sum_{m=1}^{M}a_{uu_i^c}^{m}K_m(T_c-t_i^c)\biggr)\\
&+\sum_{i=1}^{N_c}\biggl(p_{ii}\log\frac{\mu_{u_i^c}}{p_{ij}}
+\sum_{j=1}^{i-1}\sum_{m=1}^{M}p_{ij}^{m}\log\frac{a_{u_i^c u_j^c}^{m}\kappa_m(\tau_{ij}^c)}{p_{ij}^m}\biggr)
\biggr\rbrace,
\end{aligned}
\end{eqnarray*}
$p_{ii}={\mu_{u_i^c}^{(k)}}/{\lambda_{u_i^c}^{(k)}(t_i^c)}$ and $p_{ij}^m={a_{u_i^c u_j^c}^{m,(k)}\kappa_m(\tau_{ij}^c)}/{\lambda_{u_i^c}^{(k)}(t_i^c)}$. 
$\lambda_u^{(k)}(t)$ is the conditional intensity function computed with current parameters. 
When there is pairwise similarity constraint, we rewrite $E(\bm{A})$ given current parameters as 
\begin{eqnarray*}
\begin{aligned}
E_{\Theta}^{(k)}=\sideset{}{_{u=1}^{U}}\sum\sideset{}{_{u'\in\mathcal{C}_u}}\sum\|a_{u\cdot}-a_{u'\cdot}^{(k)}\|_F^2+\|a_{\cdot u'}-a_{\cdot u}^{(k)}\|_F^2.
\end{aligned}
\end{eqnarray*}

Replacing $\mathcal{L}_{\Theta}$ and $E(\bm{A})$ with $Q_{\Theta}^{(k)}$ and $E_{\Theta}^{(k)}$ respectively, we decouple parameters and obtain the surrogate objective function $F=-Q_{\Theta}^{(k)}+\alpha_{S}\|\bm{A}\|_1+\alpha_{G}\|\bm{A}\|_{1,2}+\alpha_{P}E_{\Theta}^{(k)}$.
Then, we update each individual parameter via solving $\frac{\partial F}{\partial \Theta}=\bm{0}$, and obtain the following closed form updates:
\begin{eqnarray}
\label{updateMu}
\mu_{u}^{(k+1)} = (\sideset{}{_{c=1}^{C}}\sum\sideset{}{_{u_i^c=u}}\sum p_{ii})/(\sideset{}{_{c=1}^{C}}\sum T_c),\\
\label{updateA}
a_{uu'}^{m,(k+1)}= (-B+\sqrt{B^2-4AC})/(2A),
\end{eqnarray}
\begin{eqnarray*}
\begin{aligned}
&A=\frac{\alpha_{G}}{\|a_{uu'}^{(k)}\|_2}+2(|\mathcal{C}_{u}|+|\mathcal{C}_{u'}|)\alpha_{P}',
~~\alpha_{P}'=\begin{cases}
\alpha_P, & u'\in \mathcal{C}_u\\
0, & \mbox{others}
\end{cases}\\
&B=\sideset{}{_{c=1}^{C}}\sum\sideset{}{_{u_i^c=u'}}\sum K_m(T_c-t_i^c)+\alpha_{S}\\
&\quad\quad -2\alpha_{P}'(\sideset{}{_{v\in\mathcal{C}_u}}\sum a_{vu'}^{m,(k)}+\sideset{}{_{v'\in\mathcal{C}_{u'}}}\sum a_{uv'}^{m,(k)}),\\
&C=-\sideset{}{_{c=1}^{C}}\sum\sideset{}{_{u_i^c=u}}\sum\sideset{}{_{u_j^c=u'}}\sum p_{ij}^{m}.
\end{aligned}
\end{eqnarray*}

Furthermore, for solving sparse-group-lasso (SGL), we apply the soft-thresholding method in~\cite{simon2013sparse} to shrink the updated parameters. 
Specifically, we set $a_{uu'}^{(k+1)}$ to all-zero if the following condition is holds:
\begin{eqnarray}\label{cond1}
\begin{aligned}
\|S_{\eta\alpha_{S}}(a_{uu'}^{(k+1)}-\eta\nabla_{a_{uu'}}Q|_{a_{uu'}^{(k)}} )\|_2\leq \eta\alpha_{G},
\end{aligned}
\end{eqnarray}
where $S_{\alpha}(z)=sign(z)(|z|-\alpha)_{+}$ achieves soft-thresholding for each element of input. $\nabla_{x}f|_{x_0}$ is the subgradient of function $f$ at $x_0$ w.r.t. variable $x$. 
We have $Q=-Q_{\Theta}^{(k)}+\alpha_{P}E(\bm{A})$, and
$\eta$ is a small constant. 
For the $a_{uu'}^{(k+1)}$ unsatisfying~(\ref{cond1}), we shrink it as
\begin{eqnarray}\label{update1}
\begin{aligned}
a_{uu'}^{(k+1)}=&\left(1-\frac{\eta \alpha_G}{\| S_{\eta\alpha_S}(a_{uu'}^{(k+1)}-\eta\nabla_{a_{uu'}}Q|_{a_{uu'}^{(k)}} ) \|_2} \right)_{+}\\
&\times S_{\eta\alpha_S}(a_{uu'}^{(k+1)}-\eta\nabla_{a_{uu'}}Q|_{a_{uu'}^{(k)}} ) 
\end{aligned}
\end{eqnarray}

In summary, Algorithm~\ref{algLearning} gives the scheme of our MLE-based algorithm with sparse-group-lasso and pairwise similarity constraints, which is called MLE-SGLP for short. 
The detailed derivation is given in the appendix. 
\begin{algorithm}
   \caption{Learning Hawkes Processes (MLE-SGLP)}
   \label{algLearning}
\begin{algorithmic}[1]
   \STATE \textbf{Input:} Event sequences $\mathcal{S}=\{s_c\}_{c=1}^{C}$, parameters $\alpha_S$, $\alpha_G$, (optional) clustering structure and $\alpha_P$.
   \STATE \textbf{Output:} Parameters of model, $\bm{\mu}$ and $\bm{A}$.
   \STATE Initialize $\bm{\mu}=[\mu_u]$ and $\bm{A}=[a_{uu'}^m]$ randomly.
   \REPEAT
   \REPEAT
   \STATE Update $\bm{\mu}$ and $\bm{A}$ via~(\ref{updateMu}) and~(\ref{updateA}), respectively.
   \UNTIL{convergence}
   \STATE \textbf{for}~{$u, u'=1:U$}
   \STATE \quad\textbf{if}~(\ref{cond1}) holds, $a_{uu'}=\bm{0}$; \textbf{else}, update $a_{uu'}$ via (\ref{update1}).
   \UNTIL{convergence}
\end{algorithmic}
\end{algorithm}

\subsection{Adaptive Selection of Basis Functions}
Although the nonparametric models in~\cite{lemonnier2014nonparametric,zhouke2013learning} represent impact functions as we do via a set of basis functions, they do not provide guidance for the selection process of basis functions. 
A contribution of our work is proposing a method of selecting basis functions founded on sampling theory~\cite{alan1989discrete}. 
Specifically, we focus on the impact functions satisfying following assumptions.
\begin{assum}\label{ass1}
\emph{
(i)~$\phi(t)\geq 0$, and $\int_{0}^{\infty}\phi(t)dt<\infty$. (ii)~For arbitrary $\epsilon>0$, there always exists a $\omega_0$, such that $\int_{\omega_0}^{\infty}|\hat{\phi}(\omega)| d\omega\leq \epsilon$. $\hat{\phi}(\omega)$ is the Fourier transform of $\phi(t)$.
}
\end{assum}
The assumption~(i) guarantees the existence of $\hat{\phi}(\omega)$, while the assumption~(ii) means that we can find a function with a bandlimit, denoted as $\frac{\omega_0}{2\pi}$, to approximate the target impact function with bounded residual. 
Based on these two assumptions, the representation of impact function in~(\ref{represent}) can be explained as a sampling process. 
The $\{a_{uu'}^{m}\}_{m=1}^{M}$ can be viewed as the discretized samples of $\phi_{uu'}(t)$ in $[0,T]$ and $\kappa_m(t)=\kappa_{\omega}(t,t_m)$ is sampling function (i.e., sinc or Gaussian function\footnote{For Gaussian filter $\kappa_{\omega}(t, t_m)=\exp( -(t-t_m)^2/(2\sigma^2) )$, its bandlimit is defined as $\omega = \sigma^{-1}$.}) corresponding to a low-pass filter with cut-off frequency $\omega$. 
$t_m$ is the sampling location corresponding to $a_{uu'}^m$ and the sampling rate is $\frac{\omega}{\pi}$.
The Nyquist-Shannon theorem requires us to have $\omega=\omega_0$, at least, such that the sampling rate is high enough  (i.e., $\frac{\omega_0}{\pi}$, twice bandlimit) to approximate the impact function.  
Accordingly, the number of samples is $M=\lceil\frac{T\omega_0}{\pi}\rceil$, where $\lceil x \rceil$ returns the smallest integer larger than or equal to $x$. 

Based on the above argument, the core of selecting basis functions is estimating $\omega_0$ for impact functions. 
It is hard because we cannot observe impact functions directly. 
Fortunately, based on~(\ref{intensity}) we know that the bandlimits of impact functions cannot be larger than that of conditional intensity functions $\lambda(t)=\sum_{u=1}^{U}\lambda_u(t)$. 
When sufficient training sequences $\mathcal{S}=\{s_c\}_{c=1}^{C}$ are available, we can estimate $\lambda(t)$ via a Gaussian-based kernel density estimator:   
\begin{eqnarray}\label{intensityAll}
\begin{aligned}
\lambda(t)=\sideset{}{_{c=1}^{C}}\sum\sideset{}{_{i=1}^{N_c}}\sum G_h(t-t_i^c).
\end{aligned}
\end{eqnarray}
Here $G_h(\cdot)$ is a Gaussian kernel with the bandlimit $h$. 
Applying Silverman's rule of thumb~\cite{silverman1986density}, we set optimal $h=(\frac{4\hat{\sigma}^5}{3\sum_c N_c})^{0.2}$, where $\hat{\sigma}$ is the standard deviation of time stamps $\{t_i^c\}$. 
Therefore, given the upper bound of residual $\epsilon$, we can estimate $\omega_0$ from the Fourier transformation of $\lambda(t)$, which actually does not require us to compute $\lambda(t)$ via (\ref{intensityAll}) directly. 
In summary, we propose Algorithm~\ref{algBasis} to select basis functions and more detailed analysis is given in the appendix. 
\begin{algorithm}
   \caption{Selecting basis functions}
   \label{algBasis}
\begin{algorithmic}[1]
   \STATE \textbf{Input:} $\mathcal{S}=\{s_c\}_{c=1}^{C}$, residual's upper bound $\epsilon$.
   \STATE \textbf{Output:} Basis functions $\{\kappa_{\omega_0}(t, t_m)\}_{m=1}^{M}$.
   \STATE Compute $\left(\sum_{c=1}^{C}N_c\sqrt{2\pi h^2}\right)e^{-\frac{\omega^2 h^2}{2}}$ to bound $|\hat{\lambda}(\omega)|$. 
   \STATE Find the smallest $\omega_0$ satisfying $\int_{\omega_0}^{\infty}|\hat{\lambda}(\omega)| d\omega\leq \epsilon$. 
   \STATE The proposed basis functions $\{\kappa_{\omega_0}(t, t_m)\}_{m=1}^{M}$ are selected, where $\omega_0$ is the cut-off frequency of basis function and $t_m=\frac{(m-1)T}{M}$, $M=\lceil\frac{T\omega_0}{\pi}\rceil$.
\end{algorithmic}
\end{algorithm}

\subsection{Properties of The Proposed Method}
Compared with existing state-of-art methods, e.g., the ODE-based algorithm in~\cite{zhouke2013learning} and the Least-Squares (LS) algorithm in~\cite{eichler2015graphical}, our algorithm has following advantages. 

\textbf{Computational complexity:} Given a training sequence with $N$ events, the ODE-based algorithm in~\cite{zhouke2013learning} represents impact functions by $M$ basis functions, where each basis function is discretized to $L$ points. 
It learns basis functions and coefficients via alternating optimization --- coefficients are updated via the MLE given basis functions, and then, the basis functions are updated via solving $M$ Euler-Lagrange equations. 
The complexity of the ODE-based algorithm per iteration is $\mathcal{O}(MN^3U^2+ML(NU+N^2) )$.
The LS algorithm in~\cite{eichler2015graphical} directly discretizes the timeline into $L$ small intervals. 
In such a situation, impact functions are discretized to $L$ points. 
The computational complexity of the algorithm is $\mathcal{O}(NU^3L^3)$. 
In contrast, our algorithm is based on known basis functions and does not estimate impact function via discretized points. 
The computational complexity of our algorithm per iteration is $\mathcal{O}(MN^3U^2)$. 
For getting accurate estimation, the ODE-based algorithm sampling basis functions densely. 
The LS algorithm needs to ensure that there is at most one event in each interval.
In other words, both two competitors require $L\gg N$. 
On the other hand, our algorithm converges quickly via few iterations.
Therefore, the computational complexity of the LS algorithm is the highest among the the three, and our complexity is at least comparable to that of the ODE-based algorithm.

\textbf{Convexity:} Both LS algorithm and ours are convex and can achieve global optima. 
The ODE-based algorithm, however, learns basis functions and coefficients alternatively. 
It is not convex and is prune to a local optima. 

\textbf{Inference of Granger causality:} Neither the ODE-based algorithm nor the LS algorithm considers to infer the Granger causality graph of process when learning model. 
Without suitable regularizers on impact functions, the impact functions learned by these two algorithms are non-zero generally, which cannot indicate the Granger causality graph exactly. 
What is worse, the LS algorithm even may obtain physically-meaningless impact functions with negative values.
To the best of our knowledge, our algorithm is the first attempt to solving this problem via combining MLE of the Hawkes process with sparse-group-lasso, which learns the Granger causality graph robustly, especially in the case having few training sequences.

\section{Experiments} 
For demonstrating the feasibility and the efficiency of our algorithm (MLE-SGLP), we compare it with the state-of-art methods, including the ODE-based method in~\cite{zhouke2013learning}, the Least-Squares (LS) method in~\cite{eichler2015graphical}, on both synthetic and real-world data.
We also investigate the influences of regularizers via comparing our algorithm with its variants, including the pure MLE without any regularizer (MLE), the MLE with group-lasso (MLE-GL), and the MLE with sparse regularizer (MLE-S). 
For evaluating algorithms comprehensively, given estimate $\tilde{\Theta}=\{\tilde{\bm{\mu}},\tilde{\bm{A}}\}$, we apply the following measurements:
1) The log-likelihood of testing data, \emph{Loglike};
2) the relative error of $\bm{\mu}$, $e_{\mu}=\frac{\|\tilde{\bm{\mu}}-\bm{\mu}\|_2}{\|\bm{\mu}\|_2}$;
3) the relative error of $\Phi(t)=[\phi_{uu'}(t)]$, $e_{\phi}=\frac{1}{U^2}\sum_{u,u'}\frac{\int_{0}^{T}|\tilde{\phi}_{uu'}(t)-\phi_{uu'}(t)|dt}{\int_{0}^{T}\phi_{uu'}(t)dt}$; 
4) \emph{Sparsity of impact function} --- the Granger causality graph is indicated via all-zero impact functions.

\subsection{Synthetic Data} 
We generate two synthetic data sets using sine-like impact functions and piecewise constant impact function respectively. 
Each of them contains $500$ event sequences with time length $T=50$ generated via a Hawkes process with $U=5$. 
The exogenous base intensity of each event type is uniformly sampled from $[0,\frac{1}{U}]$. 
The sine-like impact functions are generated as
\begin{eqnarray*}
\begin{aligned}
\phi_{uv}(t) = 
\begin{cases}
b_{uv}(1 - \cos(\omega_{uv} t -\pi s_{uv}) ), & t\in [0, \frac{2-s_{uv}}{4\pi \omega_{uv}}],\\
0, & \mbox{otherwise,}
\end{cases}
\end{aligned}
\end{eqnarray*}
where $\{b_{uv},\omega_{uv},s_{uv}\}$ are set as $\{0.05, 0.6\pi, 1\}$ when $u,v\in\{1,2,3\}$, $\{0.05, 0.4\pi, 0\}$ when $u,v\in\{4,5\}$, $\{0.02, 0.2\pi, 0\}$ when $u~(\mbox{or}~v)=4,~v~(\mbox{or}~u)\in\{1,2,3\}$.
The piecewise constant impact functions are the truncated results of above sine-like ones.

We test various learning algorithms on each of the two data sets with $10$ trials, respectively. 
In each trial, $C=\{50,...,250\}$ sequences are chosen randomly as training set while the rest $250$ sequences are chosen as testing set. 
In all trials, Gaussian basis functions are used, whose number and bandlimit are decided by Algorithm~\ref{algBasis}.
We test our algorithm with various parameters in a wide range, where $\alpha_P, \alpha_S, \alpha_G\in [10^{-2},10^4]$. 
According to the \emph{Loglike}, we set $\alpha_S=10$, $\alpha_G=100$, $\alpha_P=1000$. 
The \emph{Loglike}'s curves w.r.t. the parameters are shown in the appendix. 

The testing results are shown in Fig.~\ref{FigSynRes}. 
We can find that our learning algorithm performs better than other competitors on both data sets, i.e., higher \emph{Loglike}, lower $e_{\mu}$ and $e_{\phi}$, w.r.t. various $C$. 
Especially when having few training sequences, the ODE-based and the LS algorithm need to learn too many parameters from insufficient samples so they are inferior to our MLE-SGLP algorithm and its variants because of the over-fitting problem. 
By increasing the number of training sequences, the performance of the ODE-based algorithm does not improve a lot --- the nature of non-convexity may lead the ODE-based algorithm to fall into local optimal. 
All MLE-based algorithms are superior to the ODE-based algorithm and the LS algorithm, and the proposed regularizers indeed help to improve learning results of MLE. 
Specifically, if the clustering structure is available, our MLE-SGLP algorithm will obtain the best results. 
Otherwise, our MLE-SGL algorithm will be the best, which is slightly better than MLE-GL and MLE-S.

\begin{figure}[!h]
\subfigure[Sine-like case]{
\includegraphics[width=1\linewidth]{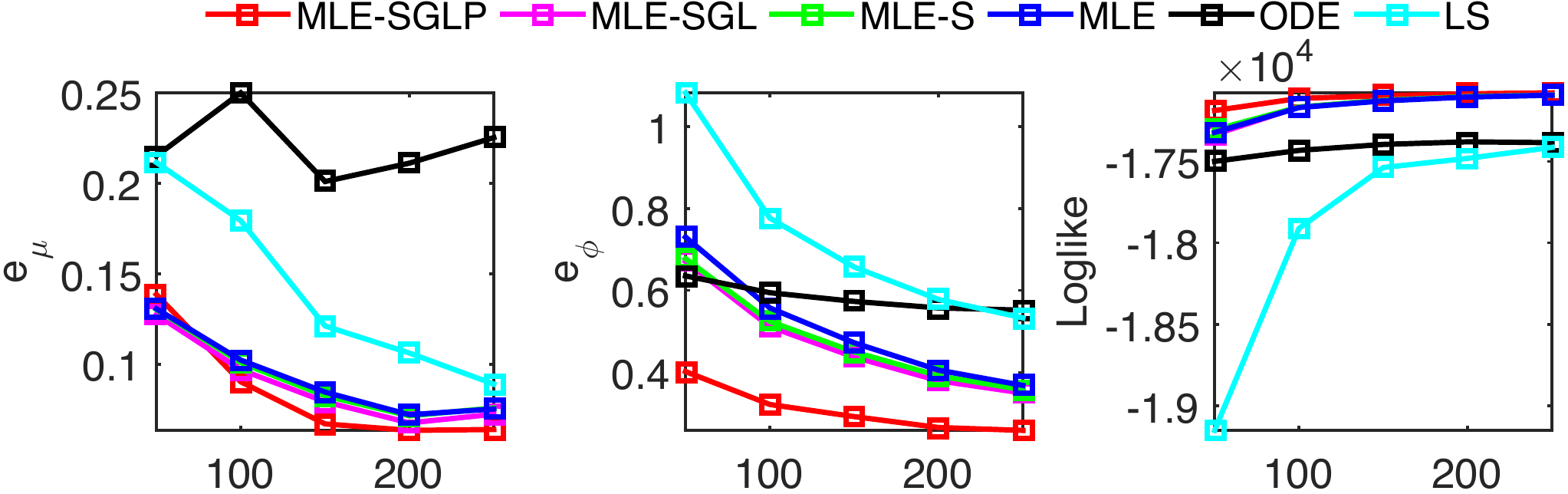}\label{FigIlluA}
}\\
\subfigure[Piecewise constant case]{
\includegraphics[width=1\linewidth]{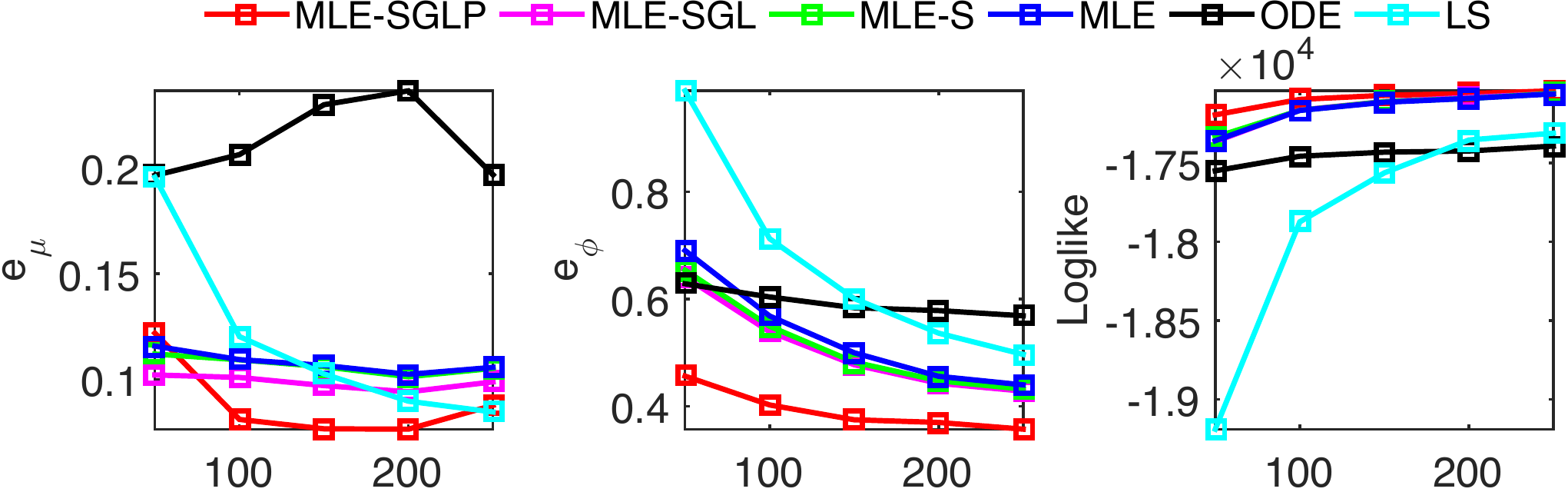}\label{FigIlluB}
}
\caption{The $e_{\mu}$, $e_{\phi}$, and \emph{Loglike} for various methods.}
\label{FigSynRes}
\end{figure}

\begin{figure}[!h]
\subfigure[Sine-like case]{
\includegraphics[width=1\linewidth]{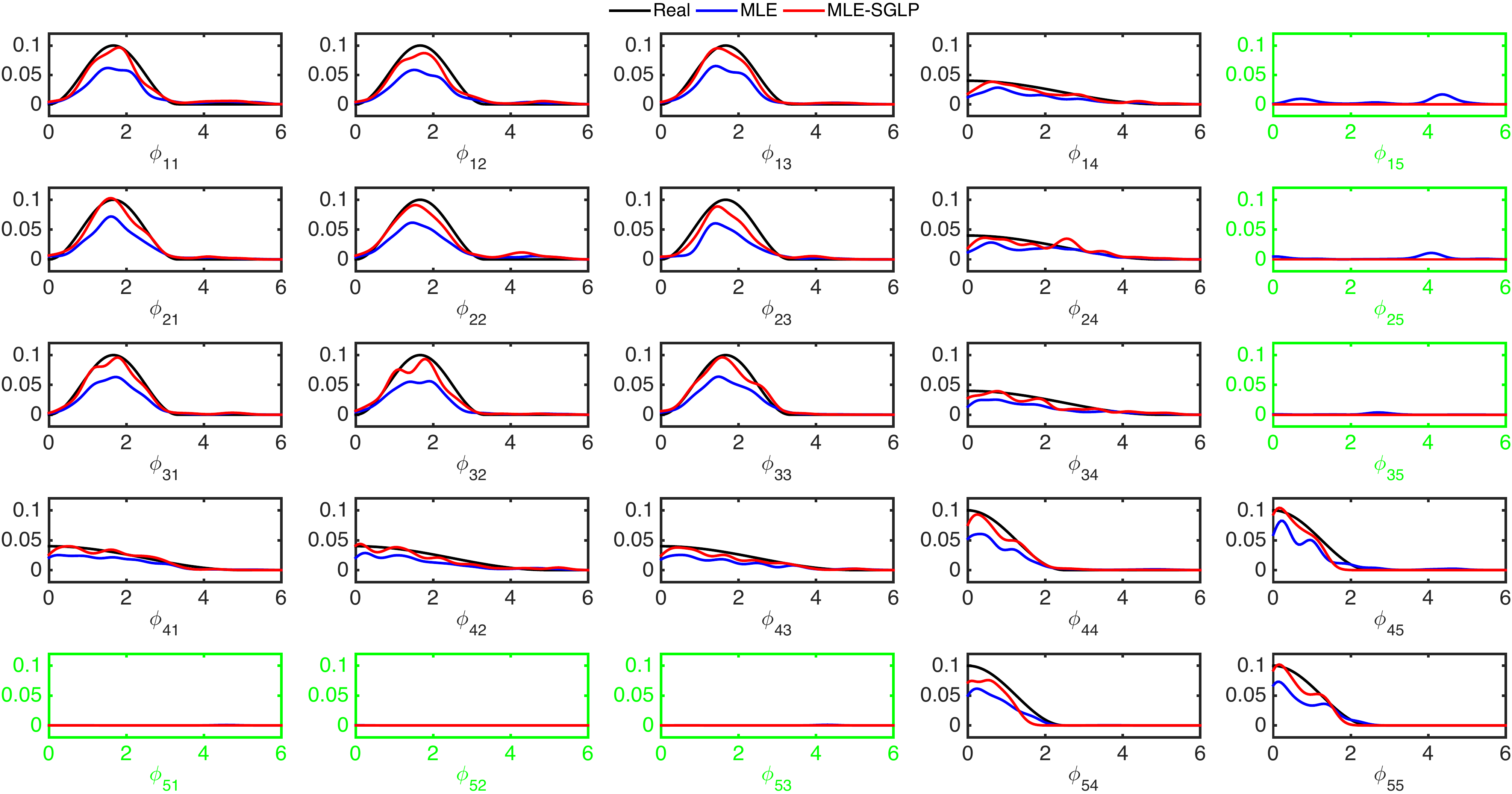}\label{FigVisA}
}\\
\subfigure[Piecewise constant case]{
\includegraphics[width=1\linewidth]{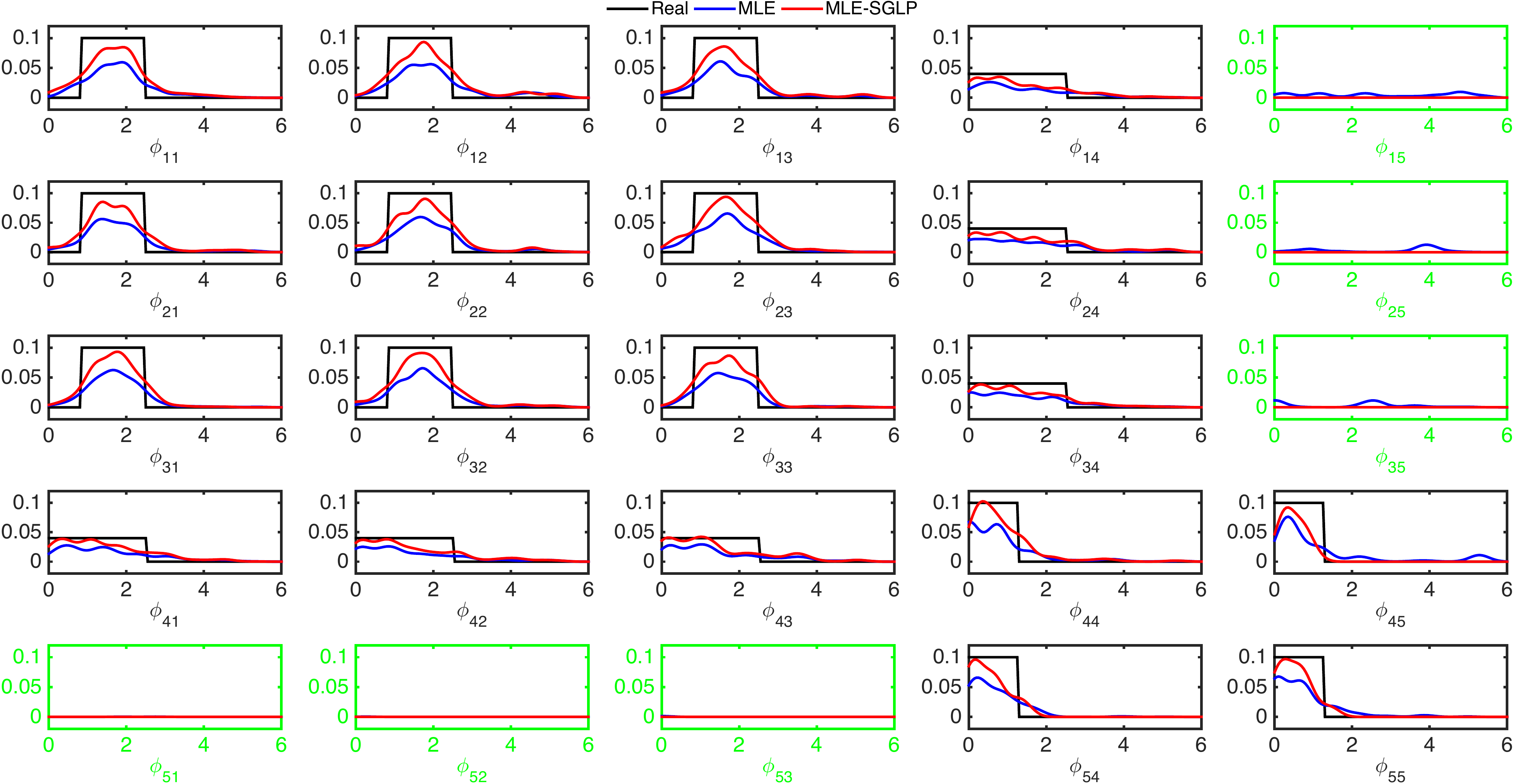}\label{FigVisB}
}
\caption{Contributions of regularizers: comparisons of impact functions obtained via MLE-SGLP and pure MLE using $500$ training sequences. The green subfigures contain the all-zero impact functions.
The black curves are real impact functions, the blue curves are the estimates from pure MLE and the red ones are proposed estimates from MLE-SGLP.}
\label{FigSynVisual}
\end{figure}

For demonstrating the importance of the sparse-group-lasso regularizer to learning Granger causality graph, Fig.~\ref{FigSynVisual} visualizes the estimates of impact functions obtained by various methods. 
The Granger causality graph of the target Hawkes process is learned by finding those all-zero impact functions (the green subfigures). 
Our MLE-SGLP algorithm obtains right all-zero impact functions while the pure MLE algorithm sometimes fails because of the lack of sparse-related regularizer. 
It means that introducing sparse-group-lasso into the framework of MLE is necessary for learning Granger causality.
Note that, even if the basis functions we select do not match well with the real case, i.e., the Gaussian basis functions are not suitable for piecewise constant impact functions, our algorithm can still learn the Granger causality graph of the Hawkes process robustly. 
As~Fig.~\ref{FigIlluB} shows, although the estimates of non-zero impact functions based on Gaussian basis functions do not fit the ground truth well, the all-zero impact functions are learned exactly via our MLE-SGLP algorithm. 

\begin{figure*}[!t]
\subfigure[Infectivity matrix]{
\includegraphics[width=0.3\linewidth]{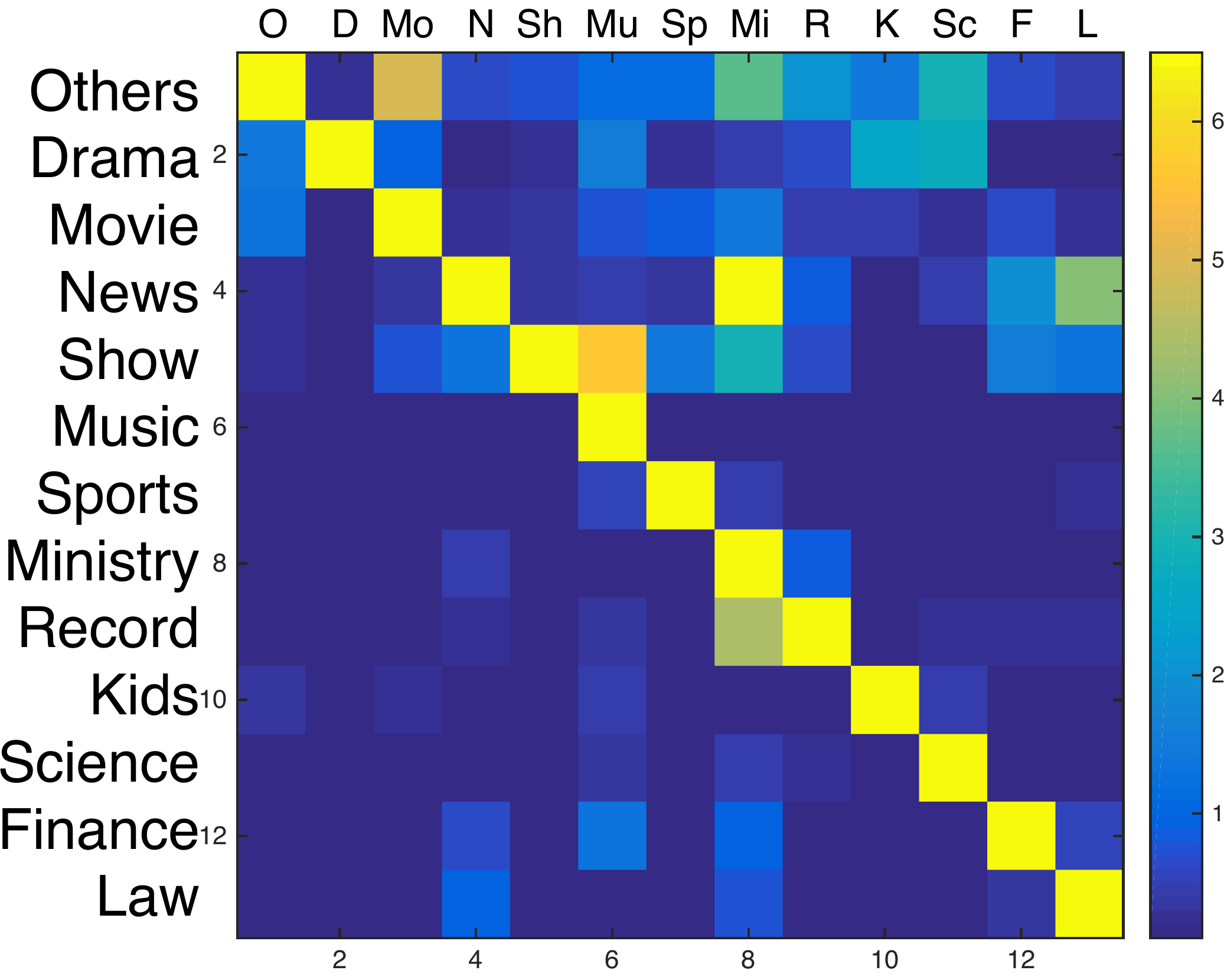}\label{figIPTVmatrix}
}
\subfigure[Top 24 impact functions]{
\includegraphics[width=0.68\linewidth]{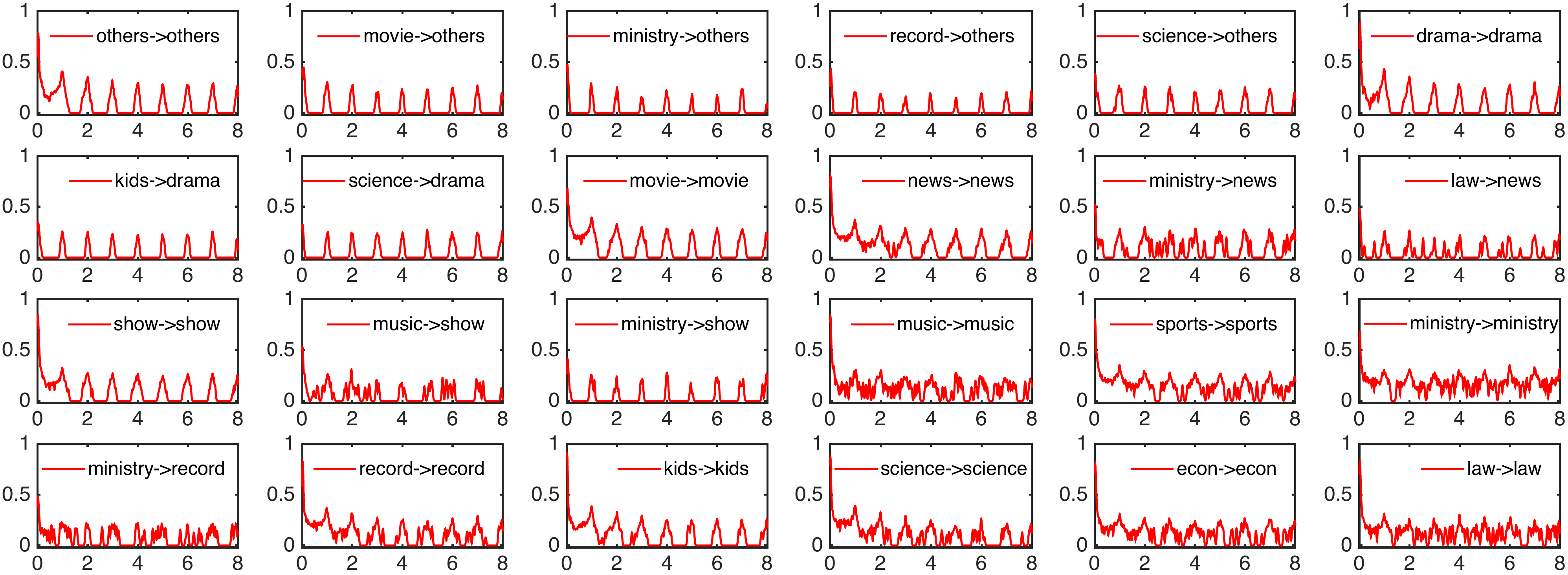}\label{figIPTVlink}
}
\caption{(a) The infectivity matrix for various TV programs. The element in the $u$-th row and the $u'$-th column is $\int_0^{\infty}\phi_{uu'}(s)ds$. 
(b) Estimates of nonzero impact functions for the IPTV data. 
By ranking the infectivity $\int_0^{\infty}\phi_{uu'}(s)ds$ from high to low, the top 24 impact functions are shown. 
For visualization, $\phi_{uu'}^{0.25}(t)$ is shown in each subfigure.}
\end{figure*}

\subsection{Real-world Data}
We test our algorithm on the IPTV viewing record data set~\cite{luo2014you,luo2015multi,luo2016learning}.
The data set records the viewing behavior of $7100$ users, i.e., what and when they watch,  in the IPTV system from January to November 2012.
$U$($=13$) categories of TV programs are predefined. 
Similar to~\cite{luo2015multi}, we model users' viewing behavior via a Hawkes process, in which the TV programs' categories exist self-and mutually-triggering patterns.
For example, viewing an episode of a drama would lead to viewing the following episodes (self-triggering) and related news of actors (mutually-triggering). 
Therefore, the causality among categories is dependent not only on the predetermined displaying schedule but also on users' viewing preferences.

We capture the Granger causality graph of programs' categories via learning impact functions. 
In this case, the pairwise sparsity is not applied because the clustering structure is not available. 
The training data is the viewing behavior in the first $10$ months and testing data is the viewing behavior in the last month. 
Considering the fact that many TV programs are daily or weekly periodic and the time length of most TV programs is about $20$-$40$ minutes, we set the time length of impact function to be $8$ days (i.e., the influence of a program will not exist over a week) and the number of samples $M=576$ (i.e., one sample per $20$ minutes). 
The cut-off frequency of sampling function is $w_0=\pi M/T$, where $T$ is the number of minutes in $8$ days. 
Table.~\ref{table1} gives \emph{Loglike} for various methods w.r.t. different training sequences. 
We can find that with the increase of training data, all the methods have improvements.
Compared with the ODE-based algorithm and pure MLE algorithm, the MLE with regularizers has better \emph{Loglike} and our MLE-SGL algorithm obtains the best result, especially when the training set is small (i.e., the sequences in one month). 
Note that here the LS algorithm doesn't work. 
Even using a PC with 16GB memory, the LS algorithm runs out-of-memory in this case because it requires to discretize long event sequences with dense samples.

\begin{table}[h]
\caption{\emph{Loglike} ($\times 10^6$) for various methods}
\label{table1}
\begin{center}
\begin{small}
\begin{sc}
\begin{tabular}{@{\hspace{2pt}}c@{\hspace{2pt}}
        @{\hspace{2pt}}c@{\hspace{2pt}}
        @{\hspace{2pt}}c@{\hspace{2pt}}
        @{\hspace{2pt}}c@{\hspace{2pt}}
        @{\hspace{2pt}}c@{\hspace{2pt}}
        @{\hspace{2pt}}c@{\hspace{2pt}}}
\hline
Alg. & ODE & MLE & MLE-S & MLE-GL & MLE-SGL\\
\hline
1 month  & -2.066 & -1.904 & -1.888 & -1.885 &\textbf{-1.880} \\
4 months & -1.992 & -1.895 & -1.880 & -1.879 &\textbf{-1.876} \\
7 months & -1.957 & -1.882 & -1.877 & -1.874 &\textbf{-1.873} \\
10 months& -1.919 & -1.876 & -1.874 & \textbf{-1.872} &\textbf{-1.872} \\
\hline
\end{tabular}
\end{sc}
\end{small}
\end{center}
\end{table}

We define the infectivity of the $u'$-th TV program category on the $u$-th one as $\int_0^{\infty}\phi_{uu'}(s)ds$, which is shown in Fig.~\ref{figIPTVmatrix}. 
It can be viewed as an adjacency matrix of the Granger causality graph. 
Additionally, by ranking the infectivity from high to low, the top 24 impact functions are selected and shown in Fig.~\ref{figIPTVlink}. 
We think our algorithm works well because the following reasonable phenomena are observed in our learning results:
 
1) All TV program categories have obvious self-triggering patterns because most of TV programs display periodically. 
Viewers are likely to watch them daily at the same time. 
Our learning results reflect these phenomena: the main diagonal elements of the infectivity matrix in~Fig.~\ref{figIPTVmatrix} are much larger than other ones, and the estimates of impact functions in~Fig.~\ref{figIPTVlink} have clear daily-periodic pattern.

2) Some popular categories having a large number of viewers and long displaying time, e.g., ``drama'', ``movie'', ``news'' and ``talk show'', are likely to be triggered by others, while the other unpopular ones having relative fewer but fixed viewers and short displaying time, e.g., ``music'', ``kids' program'', ``science'', are mainly triggered by themselves. 
It is easy to find that the infectivity matrix we learned reflects these patterns --- the non-diagonal elements involving those unpopular categories are very small or zero. 
In Fig.~\ref{figIPTVlink} the non-zero impact functions mainly involve popular categories. 
Additionally, because few viewing events about these categories are observed in the training data, the estimates of the impact functions involving unpopular categories are relatively noisy.

In summary, our algorithm performs better on the IPTV data set than other competitors. 
The learning results are reasonable and interpretable, which prove the rationality and the feasibility of our algorithm to some degree. 

\section{Conclusion}
In this paper, we learn the Granger causality of Hawkes processes according to the relationship between the Granger causality and impact functions. 
Combining the MLE with the sparse-group-lasso, we propose an effective algorithm to learn the Granger causality graph of the target process. 
We demonstrate the robustness and the rationality of our work on both synthetic and real-world data.
In the future, we plan to extend our work and analyze the Granger causality of general point processes.

\section{Acknowledgment}
This work is supported in part by NSF DMS-1317424 and NIH R01 GM108341.
Thanks reviewers for providing us with meaningful suggestions.

\section{Appendix}

\subsection{Derivation of Surrogate Objective Function}
Using the Jensen's inequality, we have following inequality for all $c$ and $i$:
\begin{eqnarray*}
\begin{aligned}
&~\log\left(\mu_{u_i^c}+\sideset{}{_{m=1}^{M}}\sum\sideset{}{_{j=1}^{i-1}}\sum a_{u_i^c u_j^c}^{m}\kappa(\tau_{ij}^c)\right)\\
\geq &
~p_{ii}\log\left(\frac{\mu_{u_i^c}}{p_{ii}}\right)+\sum_{m=1}^{M}\sum_{j=1}^{i-1}p_{ij}^m
\log\left(\frac{a_{u_i^c u_j^c}^{m}\kappa(\tau_{ij}^c)}{p_{ij}^m}\right).
\end{aligned}
\end{eqnarray*}
The equation holds if and only if $\mu_{u}=\mu_{u}^{(k)}$ and $a_{uu'}^m=a_{uu'}^{m,(k)}$. 
Therefore, we have $Q_{\Theta|\Theta^{(k)}}\geq \mathcal{L}_{\Theta}$ and $Q_{\Theta^{(k)}|\Theta^{(k)}}= \mathcal{L}_{\Theta^{(k)}}$.

\subsection{Derivation of Learning Algorithm}
We have surrogate objective function $F=-Q_{\Theta|\Theta^{(k)}}+\alpha_{S}\|\bm{A}\|_1+\alpha_{G}\|\bm{A}\|_{1,2}+\alpha_{P}E_{\Theta|\Theta^{(k)}}(\bm{A})$, where $Q=-Q_{\Theta|\Theta^{(k)}}++\alpha_{P}E_{\Theta|\Theta^{(k)}}(\bm{A})$ is the data fidelity term. 
Similar to~\cite{simon2013sparse}, we choose a group $a_{uu'}=[a_{uu'}^1,...,a_{uu'}^M]^{\top}$ to minimize and fix other parameters. 
Given current estimate $a_{uu'}^{(k)}$, we majorize $Q$ as
\begin{eqnarray}\label{major}
\begin{aligned}
Q\leq &~Q|_{a_{uu'}^{(k)}}+(a_{uu'}-a_{uu'}^{(k)})\nabla_{a_{uu'}}Q|_{a_{uu'}^{(k)}}\\
&~+\frac{1}{2\eta}\| a_{uu'}-a_{uu'}^{(k)}\|_2^2.
\end{aligned}
\end{eqnarray}
Introducing~(\ref{major}) to the surrogate objective function, we rewrite the optimization problem as
\begin{eqnarray}\label{objF}
\begin{aligned}
\min_{a_{uu'}\geq \bm{0}}&~Q|_{a_{uu'}^{(k)}}+(a_{uu'}-a_{uu'}^{(k)})\nabla_{a_{uu'}}Q|_{a_{uu'}^{(k)}}\\
&~+\frac{1}{2\eta}\| a_{uu'}-a_{uu'}^{(k)}\|_2^2++\alpha_{S}\|a_{uu'}\|_1\\
&~+\alpha_{G}\|a_{uu'}\|_{2}.
\end{aligned}
\end{eqnarray}
Because both $Q|_{a_{uu'}^{(k)}}$ and $\nabla_{a_{uu'}}Q|_{a_{uu'}^{(k)}}$ are known, we add $\frac{\eta}{2}\|\nabla_{a_{uu'}}Q|_{a_{uu'}^{(k)}}\|_2^2$ to the objective function of (\ref{objF}) and reduce $Q|_{a_{uu'}^{(k)}}$ from it, and obtain an equivalent optimization problem
\begin{eqnarray}\label{objF2}
\begin{aligned}
\min_{a_{uu'}\geq \bm{0}}&\frac{1}{2\eta}\| a_{uu'}-(a_{uu'}^{(k)}-\eta\nabla_{a_{uu'}}Q|_{a_{uu'}^{(k)}})\|_2^2\\
&+\alpha_{S}\|a_{uu'}\|_1+\alpha_{G}\|a_{uu'}\|_{2}.
\end{aligned}
\end{eqnarray}
The objective function in (\ref{objF2}) is convex, so the optimal solution is characterized by the subgradient equations.
\begin{eqnarray}
\begin{aligned}
a_{uu'}^{(k)}-\eta\nabla_{a_{uu'}}Q|_{a_{uu'}^{(k)}}-a_{uu'}=\eta\alpha_{S}\gamma+\eta\alpha_{G}\beta .
\end{aligned}
\end{eqnarray}
$\gamma=[\gamma_1,...,\gamma_M]^{\top}$, where $\gamma_m=1$ if $a_{uu'}^m>0$, and in $[0,1]$ otherwise.
$\beta = \frac{a_{uu'}}{\|a_{uu'}\|}_2$ if $a_{uu'} \neq\bm{0}$, and in the set $\{x|\|x\|_2\leq 1\}$ otherwise. 
Combining the subgradient equations with the basic algebra in~\cite{simon2013sparse}, we get that $a_{uu'}=\bm{0}$ if $\|S_{\eta\alpha_{S}}(a_{uu'}^{(k+1)}-\eta\nabla_{a_{uu'}}Q|_{a_{uu'}^{(k)}} )\|_2\leq \eta\alpha_{G}$ holds, otherwise $a_{uu'}$ satisfies
\begin{eqnarray}\label{eqQ}
\begin{aligned}
&~\left(1+\frac{\eta \alpha_G}{\|a_{uu'}\|_2}\right)a_{uu'}\\
=&~S_{\eta\alpha_S}(a_{uu'}^{(k)}-\eta\nabla_{a_{uu'}}Q|_{a_{uu'}^{(k)}}),
\end{aligned}
\end{eqnarray}
where $S_{\alpha}(z)=sign(z)(|z|-\alpha)_{+}$ achieves soft-thresholding for each element of input.
Taking the norm on both sides, $\|a_{uu'}\|_2$ can be replaced by
\begin{eqnarray}\label{norm}
\begin{aligned}
(\| S_{\eta\alpha_S}(a_{uu'}^{(k)}-\eta\nabla_{a_{uu'}}Q|_{a_{uu'}^{(k)}})\|_2-t\eta\alpha_G)_{+}.
\end{aligned}
\end{eqnarray}
Replacing the $\|a_{uu'}\|_2$ in~(\ref{eqQ}) with~(\ref{norm}), we obtain the generalized gradient step:
\begin{eqnarray}\label{update1}
\begin{aligned}
a_{uu'}^{(k+1)}=&\left(1-\frac{\eta \alpha_G}{\| S_{\eta\alpha_S}(a_{uu'}^{(k+1)}-\eta\nabla_{a_{uu'}}Q|_{a_{uu'}^{(k)}} ) \|_2} \right)_{+}\\
&\times S_{\eta\alpha_S}(a_{uu'}^{(k+1)}-\eta\nabla_{a_{uu'}}Q|_{a_{uu'}^{(k)}} ) 
\end{aligned}
\end{eqnarray} 

\subsection{Details of Basis Function Selection}
In our model, the intensity function of Hawkes process over all dimensions is:
\begin{eqnarray}\label{intensity}
\begin{aligned}
\lambda(t)&=\sum_{u=1}^{U}\lambda_u(t)\\
&=\sum_{u=1}^{U}\left(\mu_u + \sideset{}{_{u'=1}^U}\sum \int_{0}^t \phi_{uu'}(s) dN_{u'}(t-s)\right)\\
&=\sum_{u=1}^{U}\mu_u + \sum_{u=1}^{U}\sum_{t_i<t}\phi_{uu_i}(t-t_i)\\
&=\sum_{u=1}^{U}\mu_u + \sum_{u=1}^{U}\sum_{t_i<t}\sum_{m=1}^{M}a_{uu_i}^{m}\kappa_{m}(t-t_i). 
\end{aligned}
\end{eqnarray}
Applying Fourier transform, we have
\begin{eqnarray}\label{intensityF}
\begin{aligned}
\hat{\lambda}(\omega)=&\sum_{u=1}^{U}\mu_u\sqrt{2\pi} \delta(\omega) \\
&+ \sum_{u=1}^{U}\sum_{t_i<t}\sum_{m=1}^{M}a_{uu_i}^{m}e^{-j\omega t_i}\hat{\kappa}_{m}(\omega). 
\end{aligned}
\end{eqnarray}
In other words, the spectral of $\lambda(t)$ is the weighted sum of those of basis functions. 
Therefore, the cut-off frequency of basis function is bounded by that of intensity function. 

As we show in our paper, given training sequences $\mathcal{S}=\{s_c\}_{c=1}^{C}$, , where $s_c=\{(t_i^c, u_i^c)\}_{i=1}^{N_c}$, we can estimate $\lambda(t)$ empirically via a Gaussian-based kernel density estimator:  
\begin{eqnarray}\label{intensityAll}
\begin{aligned}
\lambda(t)=\sideset{}{_{c=1}^{C}}\sum\sideset{}{_{i=1}^{N_c}}\sum G_h(t-t_i^c).
\end{aligned}
\end{eqnarray}
Here $t_i^c$ is the time stamp of the $i$-th event at the $c$-th sequence. 
$G_h(t-t_i^c)=\exp(-\frac{(t-t_i^c)^2}{2h^2})$ is a Gaussian kernel with the bandwidth $h$. 

Because we only care about the selection of basis functions, we just need to estimate the spectral of $\lambda(t)$ rather than compute~(\ref{intensityAll}) directly. 
Specifically, applying Silverman's rule of thumb~\cite{silverman1986density}, we first set optimal $h=(\frac{4\hat{\sigma}^5}{3\sum_c N_c})^{0.2}$, where $\hat{\sigma}$ is the standard deviation of time stamps $\{t_i^c\}$. 
Applying Fourier transform, we compute an upper bound for the spectral of $\lambda(t)$ as
\begin{eqnarray}
\begin{aligned}
|\hat{\lambda}(\omega)| &= \left| \int_{-\infty}^{\infty}\lambda(t)e^{-j\omega t}dt\right|\\
&=\left| \sum_{c=1}^{C}\sum_{i=1}^{N_c}\int_{-\infty}^{\infty}e^{-\frac{(t-t_i^c)^2}{2h^2}}e^{-j\omega t}dt \right|\\
&\leq \sum_{c=1}^{C}\sum_{i=1}^{N_c}\left| \int_{-\infty}^{\infty}e^{-\frac{(t-t_i^c)^2}{2h^c}}e^{-j\omega t}dt \right|\\
&=\sum_{c=1}^{C}\sum_{i=1}^{N_c}\left| e^{-j\omega t_i^c}e^{-\frac{\omega^2 h^2}{2}}\sqrt{2\pi h^2} \right|\\
&\leq \sum_{c=1}^{C}\sum_{i=1}^{N_c}\left| e^{-j\omega t_i^c }\right|\left|e^{-\frac{\omega^2 h^2}{2}}\sqrt{2\pi h^2} \right|\\
&=\left(\sum_{c=1}^{C}N_c\sqrt{2\pi h^2}\right)e^{-\frac{\omega^2 h^2}{2}}.
\end{aligned}
\end{eqnarray}

Furthermore, we can compute the upper bound of the absolute sum of the spectral higher than $\omega_0$ as
\begin{eqnarray}
\begin{aligned}
&\int_{\omega_0}^{\infty}|\hat{\lambda}(\omega)|d\omega \\
\leq &\left(\sum_{c=1}^{C}N_c\sqrt{2\pi h^2}\right)\int_{\omega_0}^{\infty}e^{-\frac{\omega^2 h^2}{2}}d\omega\\
=&2\pi\left(\sum_{c=1}^{C}N_c\right)\int_{\omega_0}^{\infty}\frac{h}{\sqrt{2\pi}}e^{-\frac{\omega^2 h^2}{2}}d\omega\\
=&2\pi\left(\sum_{c=1}^{C}N_c\right)\left(\frac{1}{2}-\int_{0}^{\omega_0}\frac{h}{\sqrt{2\pi}}e^{-\frac{\omega^2 h^2}{2}}d\omega\right)\\
=&2\pi\left(\sum_{c=1}^{C}N_c\right)\left(\frac{1}{2}-\frac{1}{2}\int_{-\omega_0}^{\omega_0}\frac{h}{\sqrt{2\pi}}e^{-\frac{\omega^2 h^2}{2}}d\omega\right)\\
=&\pi\left(\sum_{c=1}^{C}N_c\right)\left(1-\frac{1}{\sqrt{2}}\mbox{erf}(\omega_0 h)\right),
\end{aligned}
\end{eqnarray}
where $\mbox{erf}(x)=\frac{1}{\sqrt{\pi}}\int_{-x}^{x}e^{-t^2}dt$. 

Therefore, give a bound of residual $\epsilon$, we can find an $\omega_0$ guaranteeing $\int_{\omega_0}^{\infty}|\hat{\lambda}(\omega)|d\omega\leq \epsilon$, or $\mbox{erf}(\omega_0 h)\geq \sqrt{2}-\frac{\sqrt{2}\epsilon}{\pi\sum_{c=1}^{C}N_c}$. 
The proposed basis functions $\{\kappa_{\omega_0}(t, t_m)\}_{m=1}^{M}$ are selected, where $\omega_0$ is the cut-off frequency of basis function and $t_m=\frac{(m-1)T}{M}$, $M=\lceil\frac{T\omega_0}{\pi}\rceil$. 

\subsection{Configuration of Parameters}
With the help of cross validation, we test our algorithm with various parameters in a wide range, where $\alpha_P, \alpha_S, \alpha_G\in [10^{-2},10^4]$. 
According to the measure \emph{Loglike}, we set $\alpha_S=10$, $\alpha_G=100$, $\alpha_P=1000$. 
The curves of \emph{Loglike} w.r.t. the three parameters are shown in the following figure. 
We can find that the learning result is relatively stable when changing the parameters in a wide range. 

\begin{figure}[!h]
\subfigure[Sine-like case]{
\includegraphics[width=1\linewidth]{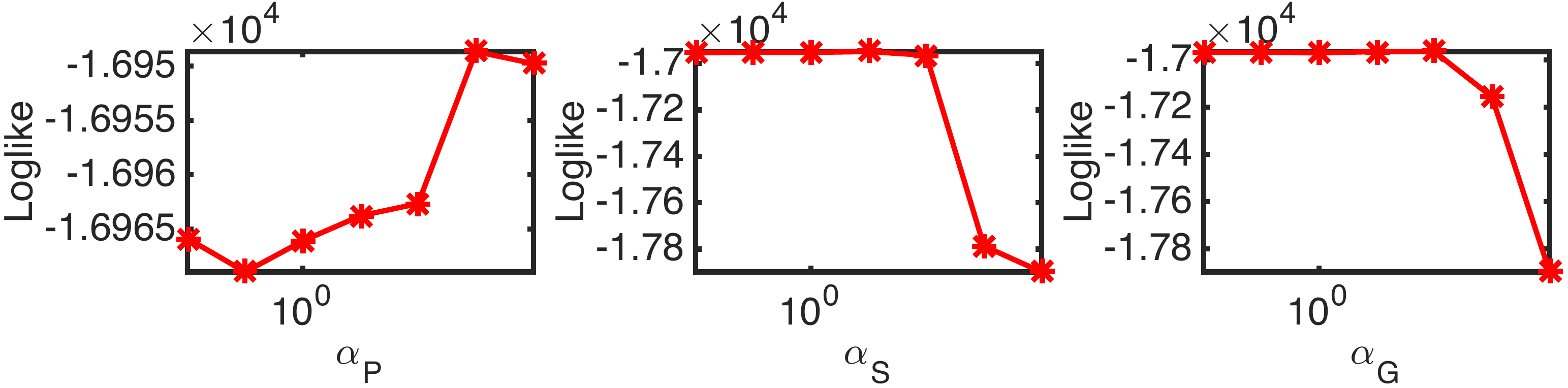}\label{FigParaA}
}\\
\subfigure[Piecewise constant case]{
\includegraphics[width=1\linewidth]{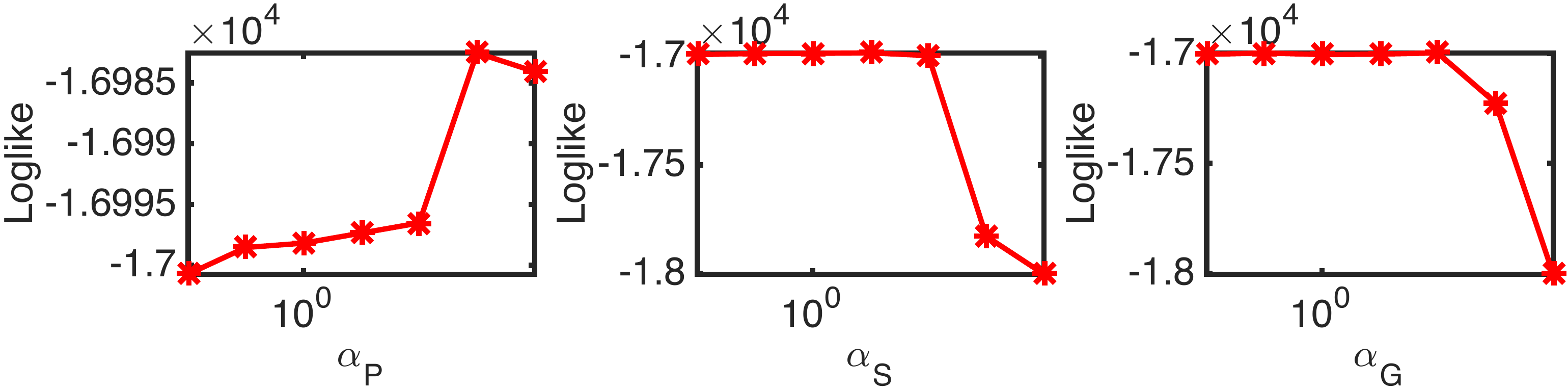}\label{FigParaB}
}
\caption{The curves of \emph{Loglike} w.r.t. the change of $\alpha_P$, $\alpha_G$ and $\alpha_S$ are shown. In each subfigure, left: $\alpha_G=100$, $\alpha_S=10$, $\alpha_P\in [10^{-2}, 10^4]$;
middle: $\alpha_G=100$, $\alpha_P=1000$, $\alpha_S\in [10^{-2}, 10^4]$;
right: $\alpha_P=1000$, $\alpha_S=10$, $\alpha_G\in [10^{-2}, 10^4]$. The number of training sequence is $250$.}\vspace{-3pt}\label{FigPara}
\end{figure}

\bibliography{example_paper}
\bibliographystyle{icml2016}

\end{document}